\documentclass[11pt]{article}
\usepackage{coling2020} 
\usepackage{times}

\usepackage{xcolor}
\usepackage{hyperref}
\usepackage{latexsym}
\usepackage{wrapfig}

\usepackage{amssymb}
\usepackage{pifont}

\usepackage{float}
\usepackage{tabularx,amsmath,booktabs}
\usepackage{pifont}

\usepackage{caption}
\usepackage[labelformat=simple]{subcaption}

\usepackage{graphicx}
\graphicspath{ {./images/} }
\usepackage{amsmath}
\usepackage{amssymb}
\usepackage{makecell}
\usepackage{multirow}
\usepackage{longtable}
\usepackage[colorinlistoftodos]{todonotes}
\usepackage{colortbl}
\usepackage{wrapfig, blindtext}
\usepackage{array}

\usepackage{tikz}
\usetikzlibrary{calc,trees,positioning,arrows,chains,shapes.geometric,%
    decorations.pathreplacing,decorations.pathmorphing,shapes,%
    matrix,shapes.symbols}

\newcommand{\fc}[0] {fact-checking}

\definecolor{OliveGreen}{RGB}{128,128,0}

\setcellgapes{8pt}

\newcolumntype{x}[1]{%
>{\centering\hspace{0pt}}p{#1}}%

\colingfinalcopy 

\title{Explainable Automated Fact-Checking: A Survey}
  
\author{Neema Kotonya \and Francesca Toni \\
  Department of Computing\\
  Imperial College London, United Kingdom\\
  \texttt{\{nk2418,ft\}@ic.ac.uk} \\}

\date{}

\begin{document}

\maketitle

\begin{abstract}
A number of exciting advances have been made in automated \fc{} thanks to increasingly larger datasets and more powerful systems, leading to improvements in the complexity of claims which can be accurately fact-checked. However, despite these advances, there are still desirable functionalities missing from the \fc{} pipeline. In this survey, we focus on the \emph{explanation} functionality 
-- that is \fc{} systems providing \emph{reasons for their predictions}. We summarize existing methods for explaining the predictions of \fc{} systems and we explore trends in this topic. 
Further, we consider what makes for good explanations in this specific domain through a comparative analysis of existing \fc{} explanations against some desirable properties.
Finally, we propose further research directions for generating \fc{} explanations, and describe how these  may lead to improvements in the research area.
\end{abstract}

\vspace{0.1cm}

\section{Introduction}
\label{sec:intro}

%
%
\blfootnote{
    
    \hspace{-0.65cm}  
    This work is licensed under a Creative Commons Attribution 4.0 International License.
    License details:
    \url{http://creativecommons.org/licenses/by/4.0/}.
}

The harms of false claims, which, through their verisimilitude, masquerade as statements of fact, have become widespread as a result of the self-publishing popularized on the modern Web 2.0, without verification. It is now widely accepted that if we are to efficiently respond to this problem, computational approaches are required \cite{fallis2015disinformation}. Falsehoods take many forms, including audio and graphical content, however a significant quantity of 
misinformation exists in natural language. For this reason, computational linguistics has a significant role to play in addressing this problem.

\begin{figure}[H]
    \centering
    \includegraphics[width=\textwidth]{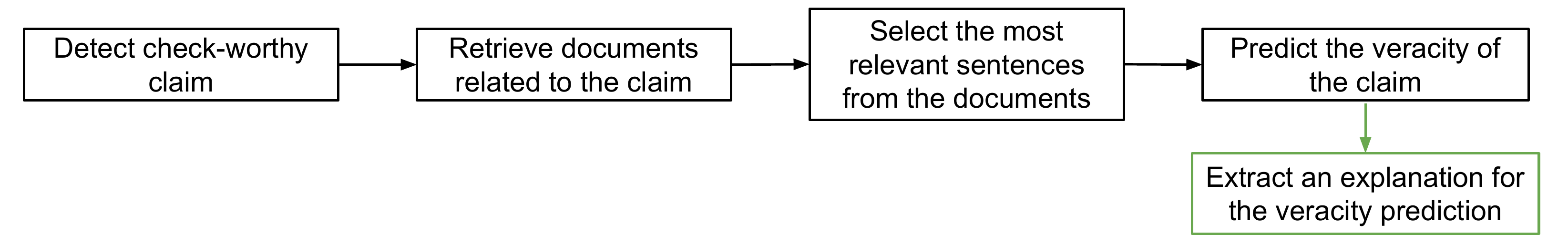}
    \caption{Pipeline for automated \fc{} with the  explanation sub-task.}
    \label{fig:pipeline}
\end{figure}

Steady progress has been made in the area of computational journalism, and more specifically automated \fc{} and its orthogonal tasks. A number of \fc{} tasks have been formulated. 
Among these tasks are the following: identifying the severity or nature of false claims, i.e., distinguishing between hoaxes, rumors and satire \cite{rubin2016fake,Rashkin:2017}; detecting check-worthy claims \cite{hassan2015detecting,hassan2017toward}; distinguishing between hyper-partisan and mainstream news \cite{Potthast:2018}; stance detection and opinion analysis in fake news \cite{Hanselowski:2018}; fauxtography, i.e., verifying the captions or claims which accompany published images \cite{Zlatkova:2019}; table-based fact verification \cite{chen2019tabfact}; credibility assessment \cite{mitra2015credbank,derczynski-etal-2017-semeval}; and an end-to-end pipeline which encompasses both the document-level evidence retrieval and sentence-level evidence selection sub-tasks of claim verification \cite{Thorne:18}. Established orthogonal tasks such as claim detection \cite{levy2014context,lippi2015context} and deception detection \cite{conroy2015automatic} also play a key role in automated \fc{}. The \fc{} pipeline is shown in Figure \ref{fig:pipeline}.

The focus of this survey is {\em explainable} automated \fc{}. Indeed, over the last few years strong progress has been made both in terms of the performance of deep learning models for \fc{} and also in terms of the availability of comprehensive datasets for training these models. 
However, not nearly as much work has been devoted to acquiring explanations for these systems, either in the form of post-hoc explanations for the outputs of \fc{} models or by hard-wiring explanation methods into these \fc{} models, even though the justification of claim verification judgments is arguably a very important part of journalistic processes when performing manual \fc{}. 

This survey has the following structure. We begin (in Section~\ref{sec:fact-checking-systems}) by laying out the current state-of-the-art in automated \fc{}. We discuss the rapid ground which has been covered in recent years with regard to both dataset construction and development of increasingly more accurate predictive models for \fc{}. We then explore (in Section~\ref{sec:exFC}) explanations for the \fc{} domain. We discuss human explanations offered by trained journalists and computationally derived \fc{} explanations. Further  (in Section~\ref{sec:analysis}), we explore limitations of the current state-of-the-art in explainable \fc{} systems.  Finally (in Section~\ref{sec:future}), we offer research directions which could be explored in order to extend this topic area and possibly mitigate some limitations we identify in the literature.

\section{Automated Fact-Checking: State of the Art}
\label{sec:fact-checking-systems}

Engineering accurate and resilient computational methods for automated \fc{} is challenging. However, steady progress has been made, and the current state-of-the-art with respect to a number of \fc{} datasets achieves promising results. 

\subsection{Datasets}

A major bottleneck associated with automated \fc{} is the shortage of available training data. However, recently there has been an increase in the volume of data available due to the rise of \fc{} websites \cite{graves2016a,lowrey2017emergence}, corpus annotation platforms, and generative language models~\cite{niewinski2019gem}. One of the first \fc{} datasets is a small corpus of 106 claims from Politifact\footnotemark[1]\footnotetext[1]{\url{https://www.politifact.com/}} \cite{Vlachos:2014}. A later dataset, EMERGENT, provides 300 claims retrieved from Twitter and Snopes\footnotemark[2]\footnotetext[2]{\url{https://www.snopes.com/}} \cite{Ferreira:2016}. 

These initial corpora for \fc{} are evidence of positive steps towards a rigorous formulation of the \fc{} task, as they introduced the task as a multi-class or graded problem \cite{Vlachos:2014} and furthermore related stance detection to claim verification \cite{Ferreira:2016}. However, due to their small size, none was sufficient for training machine learning models. Since then, a larger number of corpora have been constructed with numbers of claims which are orders of magnitude larger, i.e., in the thousands and tens of thousands, and therefore are suitable for training classifiers.

The LIAR dataset of fact-checked claims was the first to be released of a sufficient size for training machine learning models \cite{Wang:2017}. LIAR consists of 12.8K claims retrieved using the Politifact API, as well as meta-data for these claims (i.e., speaker of the claim, political affiliations of the speaker, medium through which the claim was first published). However, one element not present in the dataset the evidence used by fact-checkers to verify the claims.

Since the publication of LIAR, huge strides in progress have been made, and now there are substantially larger datasets, which also include evidence and rich with meta-data to contextualize the claims. Among these large \fc{} corpora is the manually-constructed FEVER dataset \cite{Thorne:18} which consists of 185K claims extracted from Wikipedia. There are also corpora which consist of real-world or naturally occurring claims, including claims from multiple domains, such as MultiFC \cite{Augenstein:2019}, FakeNewsNet  \cite{shu2018fakenewsnet}, and a corpus with fine-grained evidence~\cite{Hanselowski:2019}. Recently, datasets have been made available which examine \fc{} outside the politic context, e.g., in the scientific \cite{Wadden2020FactOF} and health domains \cite{kotonya-toni-2020-explainable}, which includes natural language explanations (see Table \ref{tab:data}).

\begin{table}[ht]
\begin{center}
\setcellgapes{1pt}
\makegapedcells
\begin{tabular}{lllccc}
\toprule
{\bf Dataset} &  {\bf Claims} & {\bf Sources} & {\bf Labels} & {\bf Evidence} & {\bf Explanation} \\
\midrule
Vlachos and Riedel~\shortcite{Vlachos:2014}	& 106	& Various 	&  5 &  \checkmark	& \\[0.5ex]

Mitra and Gilbert ~\shortcite{mitra2015credbank} & 1,049 & Twitter 	&  5 & 	& \\[0.5ex]
Pomerleau and Rao~\shortcite{pomerleau2017fake}  & 300 & Emergent\footnotemark[3] & 4 & \checkmark & \\[0.6ex]

Wang~\shortcite{Wang:2017}	  	& 12,836		& PolitiFact	 & 6 &  & \\[0.5ex]
P\'erez-Rosas et al.~\shortcite{perez2017automatic}& 980 & News, GossipCop & 2 &  &\\[0.5ex]
Shu et al.~\shortcite{shu2018fakenewsnet} & 23,921		& PolitiFact, GossipCop	 & 2 & \checkmark & \\[0.5ex]
Alhindi et al.~\shortcite{alhindi-etal-2018-evidence}	& 12,836		& PolitiFact	 & 6 & \checkmark &  \\[0.5ex]

Thorne et al.~\shortcite{Thorne:18}		& 185,445   	& Wikipedia			& 3	& \checkmark&  \\[0.5ex]

Augenstein et al.~\shortcite{Augenstein:2019}		& 36,534   	& Various		&  2-40	& \checkmark &   \\[0.5ex]

Hanselowski et al.~\shortcite{Hanselowski:2019}		& 6,422   	& Snopes		& 5	& \checkmark &  \\[0.5ex]

Wadden et al.~\shortcite{Wadden2020FactOF}		& 1,409   	&  S2ORC\footnotemark[4]		& 3	& \checkmark & \\[0.5ex]

Kotonya and Toni~\shortcite{kotonya-toni-2020-explainable}		& 11,832   	& Various	& 4	& \checkmark &  \checkmark \\[0.5ex]
\bottomrule

\end{tabular}
\caption{Comparison of fact-checking corpora. Vlachos and Riedel~\protect\shortcite{Vlachos:2014} take claims from Politifact and Channel 4 News, Augenstein et al.~\protect\shortcite{Augenstein:2019} extract claims from 26 \fc{} websites, and Kotonya and Toni~\protect\shortcite{kotonya-toni-2020-explainable} take claims from \fc{}, news review and news websites.}
\label{tab:data}
\end{center}
\end{table}

Decisions in dataset annotation inform formulations of the \fc{} task, e.g., the choice between binary \cite{perez2017automatic} and graded classification \cite{Vlachos:2014,Wang:2017} informs the definition
of veracity. Various annotation schemes are employed for constructing these datasets, and thus veracity labels vary considerably across datasets. For example, LIAR uses a six-point veracity system borrowed from Politifact; FEVER uses a three-way classification, with the label \textsc{Not-enough-info} to account for claims for which there is not insufficient data to arrive at a true or false judgment. The FactBank dataset \cite{sauri2009factbank} defines the factuality for claims, which are all events in the dataset, as having two dimensions: polarity (i.e., accurate, inaccurate, uncertain) and degree of certainty (i.e., certainly, probably, uncertain).

One last point to mention regarding the current available resources for \fc{} is the limitations associated with synthetic or human-crafted data for automated \fc{}. These limitations include model bias and lackluster performance on new data.
Schuster et al. \shortcite{schuster2019towards} examine the FEVER dataset, creating a symmetric dataset to show that FEVER consists of idiosyncrasies which can be exploited in order produce high accuracy \fc{} models. In order to debias the dataset, they propose a regularization method, which down weighs words and phrases which bias models.

\footnotetext[3]{\url{https://www.emergent.info}}
\footnotetext[4]{\url{https://allenai.org/data/s2orc}}

\subsection{Fact-checking Methods}

There has been steady progress in engineering accurate systems for automated \fc{}. Systems have mainly been developed in two contexts: (i) shared-tasked, focused on a single dataset, and (ii) stand-alone, 
where new datasets have been built as part of the research. 

Example  of shared tasks are the first and second  Fact Extraction and Verification (FEVER) challenges \cite{Thorne:18}. The first iteration saw twenty-four \fc{} systems submitted, with the highest performing achieving accuracy greater than 60\%. The second iteration was framed as a \textit{Build it, Break it, Fix it} problem: as well as inviting participants to build \fc{} systems, participants were also offered the opportunity to develop adversarial claims to break the builders' systems. Fixers were then invited make systems resilient to future adversarial attacks \cite{Thorne:2019}. 

Through close examination of system architectures for automated \fc{} -- both in the case of shared-tasks and standalone automated \fc{} systems -- we observe that a combination of deep neural networks (DNNs), non-DNNs and heuristic approaches are employed. The technique employed is also strongly related to the sub-task.  For example, the FEVER shared-task consists of document retrieval, evidence selection (i.e., extraction of the most pertinent sentences from the retrieved documents), relation prediction (recognizing textual entailment between claim and selected evidence sentences), and finally the implied task of label aggregation. These sub-tasks may benefit from different approaches.

Typically heuristic-based approaches are used for document retrieval. Hanselowski et al. \cite{Hanselowski:2018b} devise a document retrieval module based on TFIDF, Google Search API and named entity linking. This document retrieval module is re-used in a number of works \cite{alonso2019team,soleimani2019bert,liu2019kernel}. DNNs (e.g.  Enhanced Sequential Inference Model (ESIM) \cite{chen-etal-2017-enhanced} and decomposable attention \cite{parikh-etal-2016-decomposable}) are used for evidence selection and in some cases for joint evidence selection and relation prediction. Similar approaches are taken in the FNC1 shared-task \cite{Hanselowski:2018}, with
heuristics used for evidence selection and DNN techniques (e.g., stacked LSTMs) employed for relation prediction. The only exception is Nie et al.~\shortcite{Nie:2019}, whose method uses a DNN approach, i.e., neural semantic matching, for the entire pipeline (i.e., for all subtasks). DNN are a popular choice for the relation prediction sub-task, which is consistent with the state of the ask for this NLP task. For relation aggregation, all apart from Nie et al.~\shortcite{Nie:2019} make use of a rule-based approach or aggregate via a multilayer perception (MLP). The three \fc{} systems submitted for the Builders phase of FEVER 2.0 employ similar approaches for the document retrieval sub-task, in that they all use DNNs. However, unlike first FEVER shared-task, FEVER 2.0 models put larger emphasis on jointly learning evidence selection and relation prediction subtasks. 

Aside from the solutions developed for the FEVER shared tasks, other attempts have been made to engineer end-to-end \fc{} systems. Hassan et al.~\shortcite{hassan2017toward} develop a system to assess the veracity of claims. They source evidence from the Web and obtain knowledge bases using keyword searches. No DNN is employed in their system, instead features, e.g., part-of-speech tags and sentiment, are employed to learn a 3-way classification for claims with random forests, naive Bayes classifiers, and support vector machines. Popat et al.~\shortcite{Popat:2018} and Shu et al.~\shortcite{Shu:2019}, whose methods are also some of the first examples of explainable \fc{}, both employ DNNs for evidence selection and natural language inference. A summary of the automated \fc{} systems which are comprised of both a predictive model and explanation model is given in Table \ref{tab:transparency-systems}. In Section \ref{sec:explainable-fc-systems}, we discuss the task formulations of these explainable \fc{} in greater depth. 

\begin{table}[H]
\centering
\setcellgapes{0.75pt}
\makegapedcells
\begin{tabular}{llp{8.3cm}}
\toprule
\textbf{System}  &  \textbf{Formulation} & \textbf{Explanation methods} \\
\midrule
Popat et al.~\shortcite{Popat:2017}$^\dagger$  & Attention  & BiLSTM + CRF\\[1ex]
Popat et al.~\shortcite{Popat:2018}$^\dagger$ &  Attention  & BiLSTM + attention\\[1ex]
Shu et al.~\shortcite{Shu:2019}$^\dagger$ & Attention & BiLSTM + attention \\[1ex]
Gad-Elrab et al.~\shortcite{gad2019exfakt} & Rule discovery &  Knowledge graphs + Horn rules\\[1ex]
Ahmadi et al.~\shortcite{Ahmadi:2019}$^\dagger$ &  Rule discovery & Knowledge graphs, Horn rules +\newline probabilistic answer set programming\\[1ex]
Yang et al.~\shortcite{Yang:2019} & Attention  & CNN + self-attention\\[1ex]
Atanasova et al.~\shortcite{atanasova2020generating}$^\dagger$  & Summarization & BERT fine-tuned for extractive summarization\\[1ex]
Lu and Li.~\shortcite{lu-li-2020-gcan}$^\dagger$  & Attention & Graph-based co-attention networks\\[1ex]
Wu et al.~\shortcite{wu-etal-2020-dtca}$^\dagger$ & Attention & Co-attention self-attention networks\\[1ex]
Kotonya and Toni \shortcite{kotonya-toni-2020-explainable}  & Summarization & BERT fine-tuned for joint extractive and abstractive summarization\\[1ex]
\bottomrule
\end{tabular}

\caption{
Methods employed by explainable automated \fc{} systems. We identify where joint models are employed for prediction and explanation. $\dagger$ indicates the model performs joint veracity prediction and explanation extraction.
}
    \label{tab:transparency-systems}
\end{table}

Observing the trends, in particular the strong reliance on DNNs for textual entailment and the use of joint training for sub-tasks in the pipeline, there is evidently a trade off between system complexity and transparency. The increase in complexity of automated \fc{} methods is an even bigger incentive for acquiring explanations for these state-of-the-art systems. The observed trends also highlight that there are multiple \fc{} sub-tasks which should be considered for explanation.

\section{Explanations for \fc{}}
\label{sec:exFC}
There is increasing research interest in explanations for DNNs. These approaches include model-agnostic methods for explaining the outputs of machine learning classifiers, e.g., LIME \cite{ribeiro2016should} and SHAP \cite{NIPS2017_7062}; work on evaluating explanations \cite{narayanan2018humans}; and the formalization of desiderata (i.e., functional, operational, and user requirements) for explanations \cite{sokol2019desiderata}. 
For the purpose of this survey, we choose to examine explanations specifically with respect to the domain of automated \fc{}. To this end we will overview existing methods for explainable \fc{} and perform a comparative analysis thereof.

Before we begin our analysis, it is important to make a distinction between model \textit{interpretability} and model \textit{explainability}, which we understand as follows. 

 \paragraph{Interpretability
} 
can be understood as the ability of a machine learning model to offer a mechanism by which its decision-making can be analyzed, and possibly visualized. An interpretation is not necessarily readily understood by those who do not have specific technical expertise or else that are not well-versed in the architecture of the model, e.g., as may be the case with attention weights.  
\paragraph{Explainability
} can be understood as the ability of a machine learning model to deliver a rationale for its decisions, e.g., for a model which verifies claims through cross-checking against rules in a knowledge base, the explanation could be the full set of rules used to arrive at the final judgment. 

In this survey, we are primarily interested in natural language, human-readable explanations of machine learning models. However, at the same same time we acknowledge that model interpretations are very much a prerequisite through which human-readable explanations are achieved, but they are neither on their own sufficient nor necessarily human comprehensible.

\subsection{Fact-checking explanations by journalists}

In order to better understand explanations in the \fc{} domain, it is important that we have a background on the mechanisms of \fc{} in journalism and the explanations reported by journalists. There are a number of studies which outline an epistemology of \fc{} \cite{uscinski2013epistemology,amazeen2015revisiting,10.1111/cccr.12163}. With regard to explanations, a few points are mentioned. One is that \fc{} is typically achieved by a team of journalists who will aim to come to a consensus regarding the veracity of a claim \cite{10.1111/cccr.12163}. Graves \shortcite{10.1111/cccr.12163} makes the point that fact-checkers also work with a set of guidelines (e.g., Politifact uses the \textit{Words matter}, \textit{Context matters} and \emph{Timing} principles to form their judgments\footnotemark[5]\footnotetext[5]{\url{https://www.politifact.com/article/2013/may/31/principles-politifact/}}, see Table \ref{tab:principles}). These guidelines are analogous to explainability desiderata \cite{narayanan2018humans,sokol2019desiderata}. Graves \shortcite{graves2018understanding} states that \fc{} has three parts: \textit{identification}, \textit{verification}, and \textit{correction}. Here \textit{correction} includes flagging falsehoods and providing contextual data and fact checks, which are components of explanations.  

\begin{table}[H]
\centering
\setcellgapes{1pt}
\makegapedcells
\begin{tabular}{|p{2cm}|p{12.9cm}|}
\hline
  \textbf{Words}\newline \textbf{matter}   &  We pay close attention to the specific wording of a claim. Is it a precise statement? Does it contain mitigating words or phrases? \\
  \hline
  \textbf{Context}\newline \textbf{matters} & We examine the claim in the full context, the comments made before and after it, the question that prompted it, and the point the person was trying to make. \\
\hline
  \textbf{Timing} & Our rulings are based on when a statement was made and on the information available at that time. \\
 \hline
\end{tabular}
\caption{Three of the Politifact \fc{} principles.}
\label{tab:principles}
\end{table}

\begin{figure}[H]
\centering
\begin{subfigure}{0.495\linewidth}
\includegraphics[width=\linewidth]{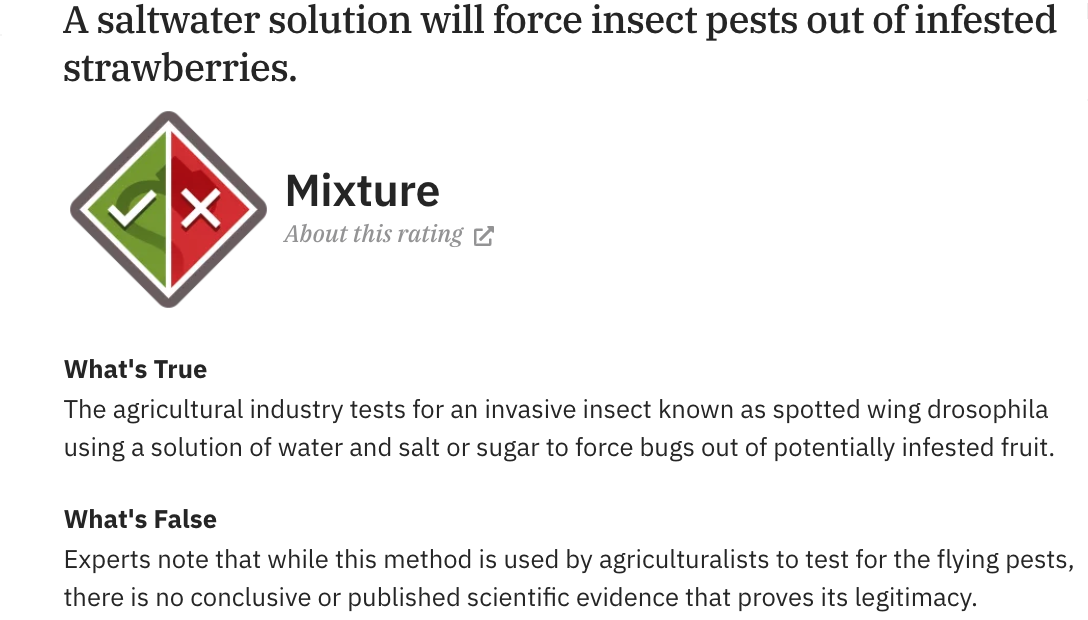}
\caption{Snopes explanation\footnotemark[6].}
\label{fig:ffs}
\end{subfigure}
\begin{subfigure}{0.495\linewidth}
\includegraphics[width=\linewidth]{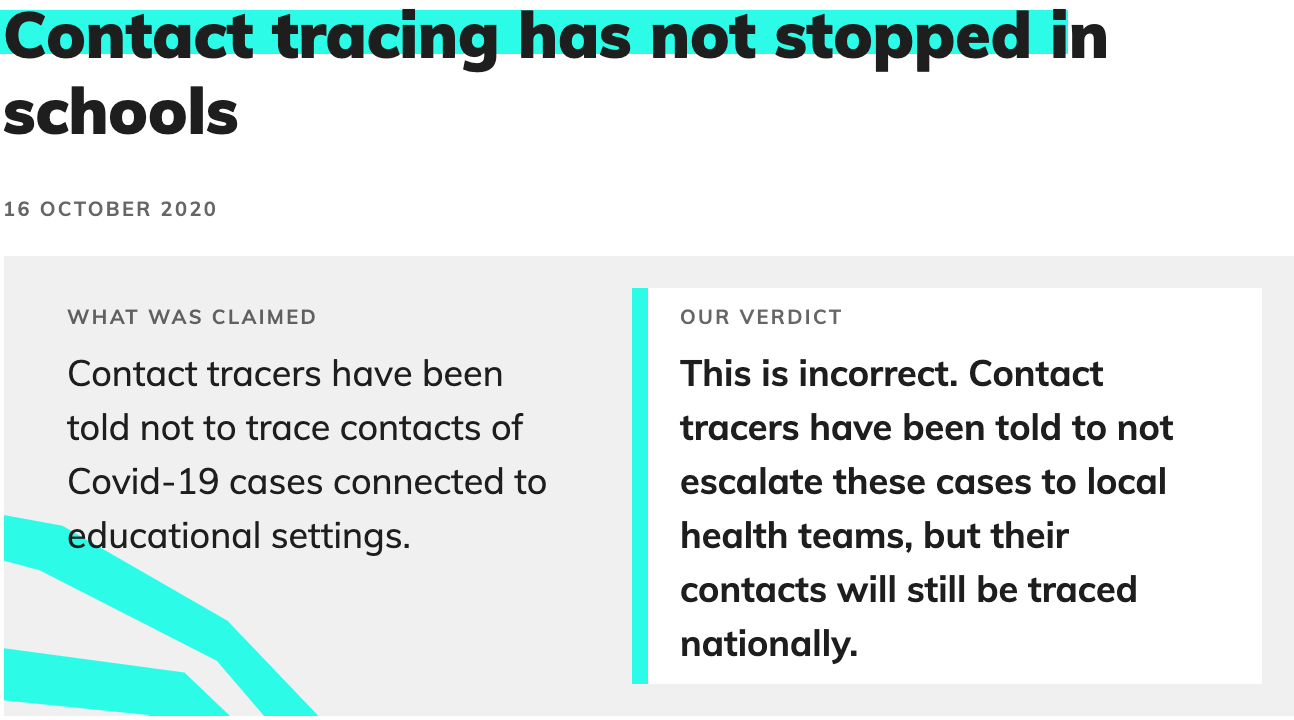}
\caption{Full Fact explanation\footnotemark[7].}
\label{fig:fullfact}
\end{subfigure}

\caption{Explanations provided by reporters on two \fc{} websites. Figure \ref{fig:journalists-explanations}\subref{fig:ffs} relates to a claim about using saltwater to remove insects from strawberries. Figure \ref{fig:journalists-explanations}\subref{fig:fullfact} is about a claim concerning contact tracing in UK schools.}
\label{fig:journalists-explanations}
\end{figure}

Another important consideration is how to gather these human explanations for \fc{}. As a certain level of expertise is required in order to produce competent \fc{} explanations, 
it would be difficult to use crowd-sourcing techniques, as has been done for acquiring human explanations for other NLP tasks \cite{Camburu:2018}. One potential source for extracting \fc{} explanations is online \fc{} platforms, as there are a considerable number of such websites \cite{graves2016a}, and it is from 
similar websites that some automated \fc{} datasets have been acquired \cite{Wang:2017,Augenstein:2019}. Figure \ref{fig:journalists-explanations} shows explanations provided by journalists on two \fc{} websites. 
Both are justifications for the veracity label attributed to the claim by the journalist, however they take two very different formats. Indeed, Figure~\ref{fig:journalists-explanations}\subref{fig:ffs} provides explanatory text which identifies \textit{what's true} and \textit{what's false} in the claim, whereas the explanation provided in Figure \ref{fig:journalists-explanations}\subref{fig:fullfact} takes the form of overall conclusion.

\footnotetext[6]{\url{https://www.snopes.com/fact-check/soaking-strawberries-in-saltwater/}}
\footnotetext[7]{\url{https://fullfact.org/health/schools-contact-tracing-not-escalate/}}

\subsection{Task Formulations}
\label{sec:explainable-fc-systems}

Progress has been made where accuracy of \fc{} systems is concerned, but a relatively small number of automated \fc{} systems have 
explainability components. Furthermore,  
various task formulations exist for generating explanations for automated \fc{} systems, e.g., explanations as text summaries. However, one aspect that the majority of methods we surveyed have in common is that their 
approaches to explainability are \textit{extractive},  i.e., the explanations produced by these systems in some way involve extracting the components of the input which are most pertinent to the prediction (see Figure \ref{fig:system-examples}). All explainable systems surveyed explain individual predictions as opposed to the \fc{} model.

\subsection{Attention-based Explanations}
These explanations take the form of some type of visualization of neural attention weights. There is a lot of interest with regards to the relationship between attention and explanation, for example the works of Jain and Wallace~\shortcite{jain-wallace-2019-attention} and Wiegreffe and Pinter~\shortcite{wiegreffe-pinter-2019-attention} take opposing views with regards to whether attention constitutes explanation.

Popat et al. \shortcite{Popat:2017}, Popat et al. \shortcite{Popat:2018}, Shu et al. \shortcite{Shu:2019},  Yang et al. \shortcite{Yang:2019}, Lu and Li~\shortcite{lu-li-2020-gcan} and Wu et al.~\shortcite{wu-etal-2020-dtca} 
 all present DNN-based methods with attention mechanisms to extract explanations. An example of explanation returned by
Popat et al. \shortcite{Popat:2018} is given in Figure \ref{fig:system-examples}\subref{fig:popat}.
Popat et al. \shortcite{Popat:2018} use a bidirectional LSTM architecture to assess the credibility of a claim with respect to a series of articles crawled from the Web. The output of the system is a credibility score for the claim based on how well is it corroborated by the articles. The trustworthiness of the claim's speaker and evidence articles, as assessed using GloVe word embeddings, are also input features used for the final output. The explanations offered by Popat et al. \shortcite{Popat:2018} are in the form of tokens highlighted in the articles, 
trained via attention. 

Shu et al.~\shortcite{Shu:2019} similarly use attention weights, 
developing a sentence-comment co-attention sub-network to highlight the most salient excerpts from the evidence articles and link them to explanatory comments offered by readers of these articles. Unlike Popat et al. \shortcite{Popat:2018}, they are able to contextualize the evidence highlighted in the articles by relating them to human assessments of those articles. However, the credibility of comments by users is not verified, as it is assumed that comment authors are trustworthy. Yang et al.~\shortcite{Yang:2019} also use, somewhat similarly, self-attention to extract n-gram explanations and linguistic analysis to extract  features, e.g., verb ratio. 

Lu and Li~\shortcite{lu-li-2020-gcan} take a slightly different approach to generating explanations. Like Shu et al.~\shortcite{Shu:2019} they make use of a co-attention mechanism, however the work, which looks at \fc{} tweets looks at explainability from three perspectives source tweets, retweet propagation, and retweeter characteristics i.e., suspicious users (see Figure \ref{fig:luli}).

Wu et al.~\shortcite{wu-etal-2020-dtca} also perform \fc{} on Twitter data. They combine decision-trees (for evidence selection), and a co-attention mechanism composed of two hierarchical self-attention networks for explainable fake news detection. The Decision Tree-based Co-Attention model (DTCA) outputs semantically relevant words from the evidence comment tweets used for \fc{}. Furthermore, the authors provide user and comment credibility scores for evidence. 

\subsection{Explanation as rule discovery}

Unlike the aforementioned works, the systems by Gad-Elrab et al. \shortcite{gad2019exfakt} and Ahmadi et al. \shortcite{Ahmadi:2019} use Horn rules and knowledge graphs to mine explanations (see Figure~\ref{fig:system-examples}\subref{fig:ahmadi}). 
A clear advantage of these systems over the attention-based explanations is that the explanations produced are more comprehensive (see Figure \ref{fig:system-examples}), as discussed later when we present a comparison of the explanations produced by each system in Table \ref{tab:comparison}. One disadvantage that these two systems have over the attention-based DNNs presented by Shu et al. ~\shortcite{Shu:2019} and Popat et al. ~\shortcite{Popat:2018} is the issue of scalability. Indeed, both systems are reliant on mining rules for each of the claim triples from a knowledge base, e.g., DBpedia, which limits the statements which can be fact-checked. However, there has been recent work which looks at alleviating this problem by creating an extensible framework for rule mining \cite{Ahmadi:2020}.

\subsection{Explanation as summarization}

\begin{figure}[ht]
\centering
\begin{subfigure}{0.495\linewidth}
    \centering
    \includegraphics[width=\linewidth]{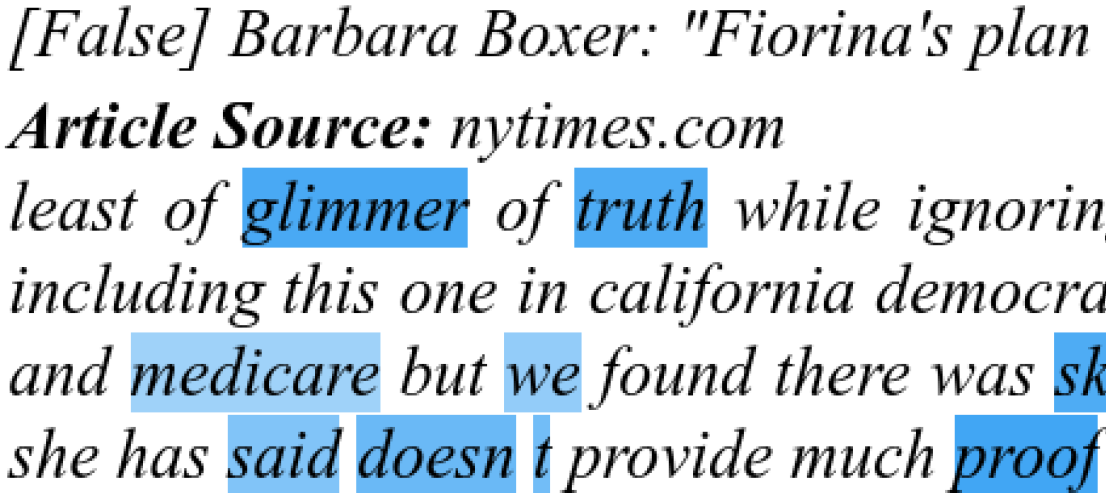}
    \caption{Attention mechanism \cite{Popat:2018}.}
    \label{fig:popat}
\end{subfigure}
\begin{subfigure}{0.495\linewidth}
    \centering
    \includegraphics[width=\linewidth]{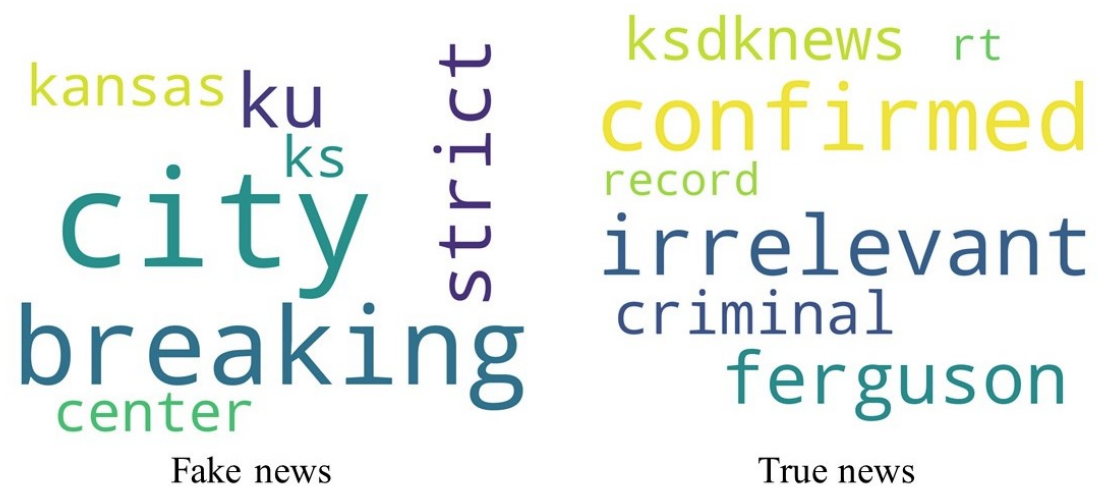}
    \caption{Attention mechanism \& user data. ~\cite{lu-li-2020-gcan}}
    \label{fig:luli}
\end{subfigure}
\begin{subfigure}{0.495\linewidth}
    \centering
    \includegraphics[width=\linewidth]{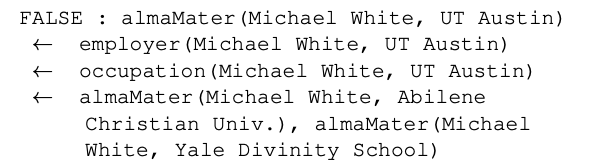}
    \caption{Rule discovery \cite{Ahmadi:2019}.}
    \label{fig:ahmadi}
\end{subfigure}
\begin{subfigure}{0.495\linewidth}
    \centering
    \includegraphics[width=\linewidth]{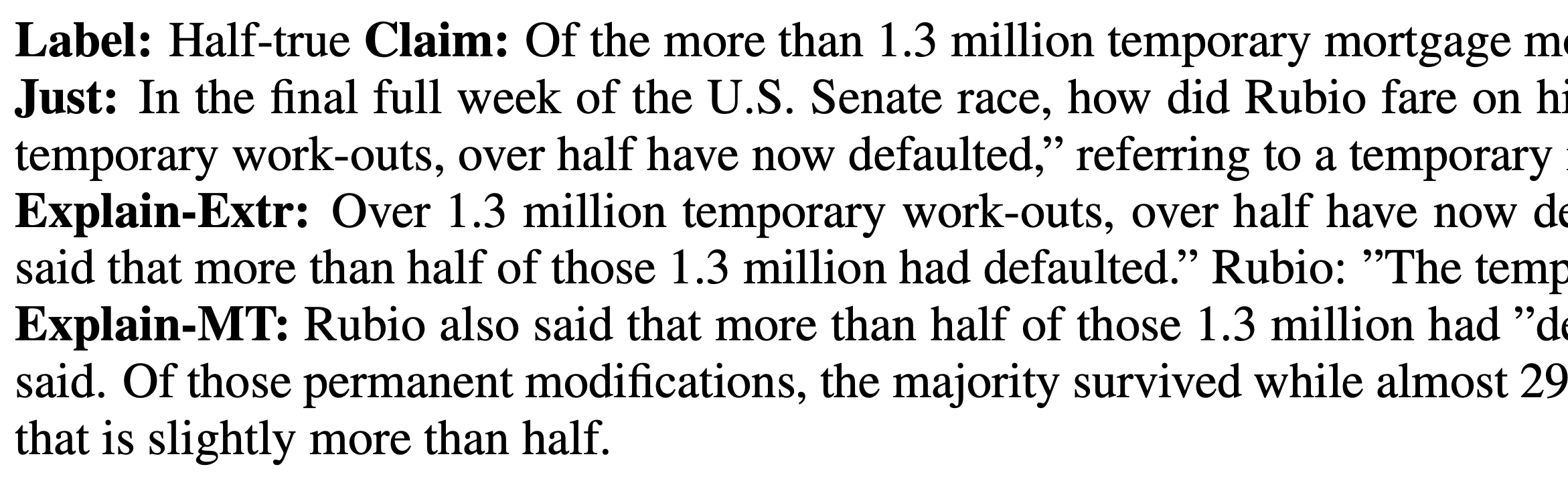}
    \caption{Text summarization \cite{atanasova2020generating}}
    \label{fig:atanasova}
\end{subfigure}
\caption{Examples of explanations generated by various task formulations.}
\label{fig:system-examples}
\end{figure}

Atanasova et al.~\shortcite{atanasova2020generating}
generate natural language explanations for \fc{}, formulating the explanation generation task as a text summarization problem  (see Figure~\ref{fig:system-examples}\subref{fig:atanasova}). Two models are trained with this in mind: one model which generates post hoc explanations, i.e., the predictive and explanation models are trained separately; and the other is jointly trained for both tasks. The jointly trained model slightly performs more poorly than the model that trains the explainer separately. The models adapt a BERT-based architecture for which the segmentation embeddings are re-purposed for text summarization \cite{liu-lapata-2019-text}. Similarly to other models we have discussed, this systems also takes an \textit{extractive} view of the explanation task, i.e., it employs extractive summarization so the outputs generated as explanations of predictions are subsets of the inputs.
Kotonya and Toni~\shortcite{kotonya-toni-2020-explainable} also model the explanation generation task as one of summarization. However, their work differs from that of Atanasova et al.~\shortcite{atanasova2020generating} in that the explanation models of the former are fine-tuned for joint extractive and abstractive summarization in order to achieve novel explanations, rather than being purely extractive summaries.

\subsection{Adversarial claims for robust \fc{}}

It is worth noting the \textit{Build it. Break it. Fix it.} theme of the FEVER 2.0 shared task \cite{thorne2019adversarial}, the \textit{Break it} component of which consisted of engineering systems for generating adversarial claims to break \fc{} systems. It is clear that adversarial examples and explanations in machine learning are orthogonal~\cite{tomsett2018failure}. The correlation between the two has been noted in machine learning from both empirical \cite{Thorne:2019} and theoretical \cite{ignatiev2019relating} stand points. In fact, for automated \fc{} there have even been attempts to generate task specific adversarial claims \cite{thorne2019adversarial}, e.g., by means of a method that uses a GPT-2 based \cite{radford2019language} model with input controls and original vocabulary \cite{niewinski2019gem}. Furthermore, a more recent work by Atanasova et al.~\shortcite{atanasova2020generatingadv} extends on this theme by generating well-formed adversarial claims, which are cohesive with the veracity label, through the use of universal adversarial triggers. These are n-grams which are inserted in the input text to confuse the model into predicting a different label for the claim.

\section{Comparative Analysis of Explainable Systems}
\label{sec:analysis}

In this section we offer a qualitative analysis of explainable \fc{} approaches. We evaluate the explainable \fc{} systems described in Section \ref{sec:explainable-fc-systems} against the eight desirable properties shown in Table \ref{tab:comparison}. Seven of these properties are borrowed from the usability and operational requirements for machine learning explanations proposed by Sokol and Flach~\shortcite{sokol2019desiderata}. We also propose one new property which is informed by the journalism literature on \fc{}. This property is explanation \emph{unbiasedness} or \emph{impartiality}, i.e., whether the \fc{} explanation attempts to highlight or mitigate bias (e.g., hyperpartisanship \cite{Potthast:2018} in the context of automated \fc{}) in the input. For this exercise, we consider whether each property has been addressed in the design of the \fc{} systems. We do not take into consideration how effectively these properties are addressed in the systems, but rather if the systems meet the minimum required for the properties.

\paragraph{ Actionable explanations.} According to Sokol and Flach~\shortcite{sokol2019desiderata}, an actionable explanation a one which users can treat as a set of guidelines indicating steps towards a desired outcome. In the case of \fc{} these would most likely take the form of factual corrections, i.e., the explanation would provide a corrected version of the erroneous claim or an  indication of which parts of the claim include misinformation. In addition, an actionable explanation could identify credible sources which counter the misinformation in the original claim. Explaining, correcting, and clarifying are complementary tasks, so naturally strides or progress towards one should aide in achieving the others. In his work on the limitations of automated \fc{}, Graves~\shortcite{graves2018understanding} identifies three main stages in automated \fc{}: identification, verification and correction. Correction should be delivered ``instantaneously, across different media, to audiences exposed to misinformation”, and consist of the following: (1) flagging repeated falsehoods, (2) providing contextual data, (3) publishing new fact checks. None of the explainable \fc{} methods we surveyed include corrections so we do not consider the explanations actionable.

\paragraph{Causal explanations.} For this property to be satisfied the explainable system would need to make use of a full causal model in order to infer causal relationships between inputs and outputted predictions. Also, such an explainable system would need to produce suitable explanations, e.g., counterfactual explanations. None of the explainable \fc{} systems we examined make use of a causal model and therefore none of them can be considered to output causal explanations.

\paragraph{Coherent explanations.} Sokol and Flach ~\shortcite{sokol2019desiderata} state that the user who is accessing the machine learning model's predictions and explanations will have prior knowledge and beliefs concerning inputs. An explanation is coherent if it is consistent with this prior knowledge. They also state that there is a high degree of subjectivity concerning this property, however, for what they describe as \textit{basic coherence}, explanations should be consistent with the natural laws. With this in mind, we have labeled the explanations which are reliant on rule-based approaches \cite{Ahmadi:2019,gad2019exfakt} as coherent because they offer an explanation in terms of rule satisfaction. Kotonya and Toni~\shortcite{kotonya-toni-2020-explainable}  define three coherence properties to evaluate explanation quality.
We do not consider explanations making use of attention-based and language generate methods to be coherent because they are non-deterministic.

\paragraph{Context-full explanations.} According to Sokol and Flach~\shortcite{sokol2019desiderata}, the \textit{contextfullness} property is satisfied by an explanation if this is presented in a way that its context can be fully understood. We decided that all the explanations generated fulfill this criterion because they are all presented in the context of larger inputs i.e., the claim and sometimes user context \cite{Shu:2019,wu-etal-2020-dtca}.

\paragraph{Interactive explanations.} Explanations which are interactive allow users to provide feedback to the system. Only two of the explainable \fc{} systems in the literature satisfy this property: CredEye \cite{Popat:2018} and XFake \cite{Yang:2019}. The CredEye systems allows users to rate (a binary \textit{yes} or \textit{no}) whether they agree with the \fc{} assessment of the system. XFake offers a visualization of various assessments of the \fc{}, e.g., linguistic analysis so that the user can understand the misinformation at different levels of granularity.

\paragraph{Unbiased or impartial explanations.} In the context of automated \fc{}, an unbiased explanation is one for which biases in the evidence data and other inputs have not influenced the explanation. Typically, for \fc{}, bias is in the form of hyper-partisan language or opinions presented as evidence. None of the systems which we examine explicitly consider this property. However we consider the rule-based approaches to be a step towards this because they output triples and not text sequences. 

\paragraph{Parsimonious explanations.} This relates to brevity, i.e., an explanation should communicate all information needed with minimal redundant text. We consider the explanations which output rules, tokens, and word clouds to be parsimonious. Atanasova et al.~\shortcite{atanasova2020generating} evaluate explanations for non-redundancy.

\paragraph{Chronological explanations.} This property is related to the time-sensitive nature of \fc{} as described by the \textit{timing} principle in Table \ref{tab:principles}. None of the systems considers this property.

\begin{table}[ht]
\centering
\setcellgapes{1.5pt}
\makegapedcells
\begin{tabular}{lcccccccc}
\toprule
\textbf{System} & \textbf{ACT} & \textbf{CSL} & \textbf{CHR} & \textbf{CNT} & \textbf{INT} & \textbf{UNB} & \textbf{PRS} & \textbf{CHN} \\[1.5ex]

\midrule
Popat et al.~\shortcite{Popat:2017}  &  &  &  & \checkmark & \checkmark &  & \checkmark & \\
Popat et al.~\shortcite{Popat:2018}  &  & &  & \checkmark &  &  & \checkmark & \\
Shu et al.~\shortcite{Shu:2019}  &  &  &   & \checkmark &  & & & \\
Gad-Elrab et al.~\shortcite{gad2019exfakt}  &  &  & \checkmark & \checkmark &  & \checkmark & \checkmark & \\
Ahmadi et al.~\shortcite{Ahmadi:2019}  &  &  & \checkmark & \checkmark &  & \checkmark &  \checkmark & \\[1ex]
Yang et al.~\shortcite{Yang:2019}  &  &  &  & \checkmark &  \checkmark &  & &\\[1ex]
Atanasova et al.~\shortcite{atanasova2020generating}  &  &  &  & \checkmark &  &  & \checkmark &\\[1ex] 
Lu and Li~\shortcite{lu-li-2020-gcan}  &  &  &  & \checkmark &  &  &  \checkmark &\\[1ex]
Wu et al.~\shortcite{wu-etal-2020-dtca}  &  &  &  & \checkmark &  &  & \checkmark &\\[1ex] 
Kotonya and Toni~\shortcite{kotonya-toni-2020-explainable}  &  &  & \checkmark & \checkmark &  &  & &\\[1ex]
\bottomrule
\end{tabular}
\caption{Properties of explanations, which are actionability (ACT), causality (CSL), coherence (CHR), contextfullness (CNT), interactiveness (INT), unbiasedness or impartiality (UNB), parsimony (PRS) and chronology or timing (CHN). \checkmark indicates a property is considered. There is no explainable \fc{} system which considers the actionability, causality or chronology properties.}
\label{tab:comparison}
\end{table}
\vspace{-0.2cm}

\section{Future Directions}
\label{sec:future}

Here we present a number of limitations in the state-of-the-art approaches to explainable \fc{} (in addition to the limitations emerging from Table~\ref{tab:comparison}). With each shortcoming we also present possible 
approaches which might be explored in order to overcome the limitation.

\paragraph{Unverifiable Claims.}

In \fc{}, there is often an issue which arises with unverifiable claims. Indeed, there is sometimes insufficient evidence to either verify or refute claims. This problem arises in both manual and computational \fc{} \cite{uscinski2013epistemology,Thorne:18}. For this reason, most formulations of the \fc{} task take this scenario into account with the inclusion of an additional class, e.g., the \textsc{not-enough-info} label in the FEVER dataset \cite{Thorne:18}.
In such cases it would be necessary to offer an explanation as to why the claim could neither be verified nor refuted or what supporting evidence was missing such as to prevent a verdict being reached (i.e., a counterfactual explanation). For many explainable systems this is particularly tricky because explanations are typically generated with respect to the inputs. The knowledge graph-based systems \cite{gad2019exfakt,Ahmadi:2019} take steps towards this, as their output 
amounts to rules that serve as both evidence and counter-evidence, but more human understandable explanations are needed which state explicitly what evidence, if any, is available to verify the claim and why it is insufficient.

\paragraph{Predictions vs. Models.}

One limitation of current explainable \fc{} systems is the emphasis on outcome-driven explanations. These typically amount to explaining the predictions with reference to relation prediction. There is no system which offers process-driven or model explanations for multiple or all sub-tasks in the \fc{} pipeline, including justifications which might shed light on why and how decisions made in earlier sub-tasks affect outputs of later sub-tasks, e.g., evidence selection and natural language inference (NLI). There is existing work on explaining NLI \cite{Thorne:2019}, with, in particular, the e-SNLI dataset containing explanation annotations for NLI \cite{Camburu:2018}. Building upon these works in explainable NLI, there are a number of directions which we envision as future avenues of research in explainable \fc{} which could result in valuable gains.

\paragraph{Evaluating explanations.}
As well as extracting explanations for predictions it is important that metrics are established for evaluating these explanations. The explainable \fc{} systems surveyed make use of both computational and human evaluations for  explanation quality. Shu et al. \shortcite{Shu:2019}, Gad-Elrab et al. \shortcite{gad2019exfakt}, Atanasova et al. \shortcite{atanasova2020generating}, Kotonya and Toni~\shortcite{kotonya-toni-2020-explainable}, and Yang et al. \shortcite{Yang:2019} employ human annotators to assess the quality of explanations generated by their models. Yang et al. \shortcite{Yang:2019} 
conduct their human evaluation on the largest number of annotators, this is the only work which considers age, gender and education level as factors when choosing participants for the evaluation study. 
Kotonya and Toni~\shortcite{kotonya-toni-2020-explainable} formalize three coherence properties for evaluating the quality of explanations: local coherence, and strong and weak global coherence. ExpClaim \cite{Ahmadi:2019} is evaluated against and outperforms CredEye \cite{Popat:2018}, and GCAN \cite{lu-li-2020-gcan} is shown to outperform dEFEND \cite{Shu:2019}. However, these evaluations are with respect to the prediction, not explanations. No evaluation is performed on explanations generated for some systems \cite{Popat:2017,Popat:2018}. In order to compare performance across explainable systems, it would be helpful to establish task-specific metrics similar to those which exist for other NLP tasks, e.g., machine translation \cite{papineni2002bleu} and text summarization \cite{lin2004rouge}.

\section{Conclusion}

In conclusion, the emerging work on explainable machine learning in the \fc{} domain shows a great deal of promise despite the particularly challenging nature of the problem. However, there are some limitations of current methods and task formulations which we highlight in this survey: all existing methods only look to explain one component of the \fc{} pipeline (relation prediction or entailment) and the systems only explain the predictions. In order to garner the most relevant, useful and insightful \fc{} explanations, a holistic and journalistic-informed approach should be taken. We identify possible directions for future research, which will hopefully extend the work achieved so far.


\bibliographystyle{coling}
\bibliography{coling2020}

\end{document}